\documentclass[conference]{IEEEtran}
\IEEEoverridecommandlockouts
\usepackage{hyperref}  
\usepackage{cite}
\usepackage{amsmath,amssymb,amsfonts}
\usepackage{algorithmic}
\usepackage{graphicx}
\usepackage{textcomp}
\usepackage{xcolor}
\usepackage{graphicx}
\usepackage{amsmath}
\usepackage{multirow}
\usepackage{amsmath}
\usepackage{multirow}
\usepackage{array}
\usepackage{booktabs}
\usepackage{array}
\usepackage{algorithm}
\usepackage{algorithmic}
\usepackage{amsmath}
\usepackage{float}
\usepackage{siunitx}
\usepackage{todonotes}
\usepackage{comment}  
\usepackage{titlesec}
\titlespacing*{\paragraph}{0pt}{0pt}{1em}

\def\BibTeX{{\rm B\kern-.05em{\sc i\kern-.025em b}\kern-.08em
    T\kern-.1667em\lower.7ex\hbox{E}\kern-.125emX}}
\begin{document}

\title{Efficient Deployment of Spiking Neural Networks on SpiNNaker2 for DVS Gesture Recognition Using Neuromorphic Intermediate Representation \\

}

\author{\IEEEauthorblockN{
Sirine Arfa\IEEEauthorrefmark{1}\IEEEauthorrefmark{2},
Bernhard Vogginger\IEEEauthorrefmark{1},
Chen Liu\IEEEauthorrefmark{1},
Johannes Partzsch\IEEEauthorrefmark{1},
Mark Sch\"{o}ne\IEEEauthorrefmark{1}
Christian Mayr\IEEEauthorrefmark{1}\IEEEauthorrefmark{6}\IEEEauthorrefmark{4}
}
\IEEEauthorblockA{\IEEEauthorrefmark{1}Chair of Highly-Parallel VLSI-Systems and Neuro-Microelectronics, Technische Universität Dresden, Germany}

\IEEEauthorblockA{\IEEEauthorrefmark{6}ScaDS.AI Dresden/Leipzig, Germany}

\IEEEauthorblockA{\IEEEauthorrefmark{4}Centre for Tactile Internet with
Human-in-the-Loop (CeTI)}

\IEEEauthorblockA{\IEEEauthorrefmark{2}Email: sirine.arfa@tu-dresden.de}


}

\maketitle

\begin{abstract}
Spiking Neural Networks (SNNs) are highly energy-efficient during inference, making them particularly suitable for deployment on neuromorphic hardware. Their ability to process event-driven inputs, such as data from dynamic vision sensors (DVS), further enhances their applicability to edge computing tasks. However, the resource constraints of edge hardware necessitate techniques like weight quantization, which reduce the memory footprint of SNNs while preserving accuracy. Despite its importance, existing quantization methods typically focus on synaptic weights quantization without taking account of other critical parameters, such as scaling neuron firing thresholds.

To address this limitation, we present the first benchmark for the DVS gesture recognition task using SNNs optimized for the many-core neuromorphic chip SpiNNaker2. Our study evaluates two quantization pipelines for fixed-point computations. The first approach employs post training quantization (PTQ) with percentile-based threshold scaling, while the second uses quantization aware training (QAT) with adaptive threshold scaling. Both methods achieve accurate 8-bit on-chip inference, closely approximating 32-bit floating-point performance. Additionally, our baseline SNNs perform competitively against previously reported results without specialized techniques. These models are deployed on SpiNNaker2 using the neuromorphic intermediate representation (NIR). Ultimately, we achieve 94.13\% classification accuracy on-chip, demonstrating the SpiNNaker2's potential for efficient, low-energy neuromorphic computing.

\end{abstract}


\section{Introduction}
Spiking neural networks (SNNs) have found applications in a wide range of areas, from image classification to gesture recognition \cite{pfeiffer2018deep}. High accuracy in SNNs is often achieved by designing large networks, which can represent more complex features than smaller models. However, this complexity makes it challenging to deploy large SNNs in resource-constrained environments such as neuromorphic chips.

Optimization methods have been introduced to address these challenges. A common approach is quantization, which lowers data precision to reduce memory usage, power consumption, and computational demands. Quantization must be carefully managed, though, as reducing precision too much can impact the network's accuracy. Two main quantization techniques are widely used: quantization aware training (QAT) and post training quantization (PTQ). 

QAT involves re-training the model with lower precision to minimize accuracy loss, making it well-suited for cases where accuracy is critical but time and data for re-training are available \cite{weng2021neural}. In contrast, PTQ does not require model re-training, making it faster and ideal for scenarios with limited training data, though it can lead to greater accuracy loss at lower bitwidth. Techniques like neuron elimination \cite{vidya2020fspinn}, weight pruning \cite{rathi2018stdp}, and stochastic neuron operations \cite{sen2017approximate} are also used to reduce the overall operations in SNNs, contributing to more efficient deployment. These combined approaches enable SNNs to perform effectively across various hardware setups while balancing accuracy and efficiency requirements.
\begin{figure*}[!htb]
    \centering
    \includegraphics[width=\textwidth]{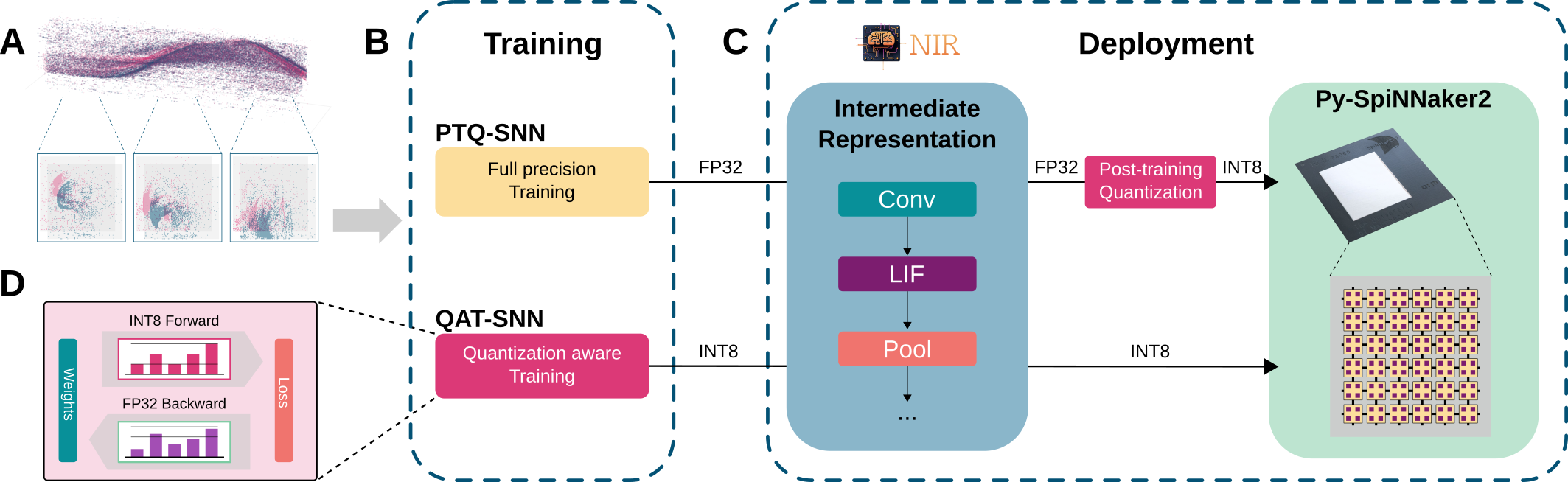}
    \caption{
        \textbf{A:} Event stream from the DVS Gestures dataset~\cite{amir2017low}
        \textbf{B:} Full precision and quantization aware training (QAT) is conducted in snnTorch~\cite{eshraghian2023training}.
        \textbf{C:} The trained models are parsed with NIR~\cite{pedersen2024neuromorphic} and mapped to SpiNNaker2 with the \textsc{py-spinnaker2} library~\cite{vogginger2023pyspinnaker2}. Post training quantization (PTQ) is applied to the full precision model.
        \textbf{D:} Quantization aware training stores as full precision representation of the weights. The forward pass converts the weights to 8-bit integers, and the backward pass updates the full precision parameters directly.
    }
    \label{fig:full-pipeline}
\end{figure*}
While weight quantization is widely studied in artificial neural networks (ANNs), its use in SNNs is relatively less explored, with only a few studies adopting quantized SNNs (QSNNs) \cite{vidya2021q}. For example, Amir et al. \cite{amir2017low} applied deterministic rounding to quantize SNN weights on the TrueNorth chip, while Eshraghian and Lu \cite{eshraghian2022fine} achieved binary weights by modifying neuron firing thresholds. Other research from Putra and Shafique \cite{vidya2021q} introduced a framework that combines PTQ and QAT using techniques of truncation and rounding. Additionally, a custom QAT framework, designed for Intel's Loihi chip, was developed based on a bit-equivalent forward pass for weights and neuron states during training \cite{lava-dl}.

Gesture recognition is a prominent application area for SNNs, particularly on neuromorphic hardware. On Intel’s Loihi chip, for instance, a gesture classification task reached \SI{89.64}{\percent} accuracy by transforming a deep neural network (DNN) into an SNN \cite{massa2020efficient}. Similarly, a CNN-based model on the TrueNorth platform achieved \SI{94.59}{\percent} accuracy on the DVS gesture dataset \cite{amir2017low}.

In this paper, we present two compact SNN models tailored for efficient gesture recognition on the DVS gesture dataset. These models are referred to as \textbf{\underline{P}-SNN}, trained with full precision and deployed on the SpiNNaker2 chip using the \underline{\textbf{P}}TQ pipeline, and \textbf{\underline{Q}-SNN}, deployed on the chip using the \underline{\textbf{Q}}AT pipeline. Both models achieve significant memory savings while delivering state-of-the-art accuracy, surpassing several state-of-the-art methods. Implemented on the SpiNNaker2 neuromorphic platform, our approach harnesses the event-driven nature of SNNs for energy-efficient processing. This design achieves high execution efficiency, benefiting from SpiNNaker2’s optimized handling of sparse, event-based data.
\newline
The main contributions of this paper are:
\begin{itemize}
    \item \textbf{End-to-End Deployment Pipeline:} We present an end-to-end pipeline for deploying deep SNN models on the SpiNNaker2 chip as shown in Figure\ref{fig:full-pipeline}. This pipeline includes model training, 8-bit quantization, conversion to the neuromorphic intermediate representation (NIR)\cite{pedersen2024neuromorphic}, and final deployment on SpiNNaker2.
    This also includes providing NIR support for quantized weight layers as a key contribution.
    \item \textbf{Comparison of Quantization Schemes:} We explore two quantization schemes for SNNs --- PTQ with percentile-based threshold scaling and QAT with adaptive threshold scaling. We analyze these methods for our use case, offering insights on their strengths and limitations. Our final pipeline serves as a roadmap for future SNN deployments on SpiNNaker2, providing both a foundation for more advanced architectures and a summary of \textit{lessons learned} during implementation.
\end{itemize}
Finally, we release our code at \url{https://gitlab.com/Sirine_Arfa/deep-snn-deployment-on-spinnaker2-single-chip-using-nir.git} to enable researchers to use our trained models and NIR graphs for DVS gesture recognition, supporting intuitive inference on other neuromorphic hardware and benchmarks.

\section{Background}
\subsection{Spiking Neuron Model and QSNNs}
The spiking neuron model adopted is the leaky integrate-
and-fire neuron \cite{eshraghian2022navigating}.
In SNNs, there are two key non-differentiable operations: spike generation and the quantization applied to weights during QAT. To handle the non-differentiability of spikes, a step function is used during the forward pass, with a surrogate gradient applied during the backward pass. 

Alternatively, an approach developed by William Severa \cite{severa2018whetstone}, known as the Whetstone method, gradually sharpens the slope of the neuron’s transfer function during training so that it asymptotically approaches a threshold function. This allows conventional training methods to be applied while ensuring the final network operates with discrete spike-based activations.

Similarly, for QAT, full-precision representations of the weights are stored, while quantized 8-bit weights are used during the forward pass as shown in Figure \ref{fig:full-pipeline}-D. Gradients are calculated during the backward pass by ignoring the non-differentiable quantization operator, and updates are applied directly to the full-precision weights. To prevent the quantization operation from nullifying the gradient, a straight-through estimator (STE) \cite{venkatesh2024squat} is used. Precisely, the surrogate gradient of the quantized weights \( w_q \) in relation to the real weights \( w_r \) is \cite{bengio2013estimating}-\cite{fan2022training}: 
\begin{equation}
\frac{\partial w_q}{\partial w_r} = 1
\end{equation}

Thus smoothing the thresholding function during training.
Alternatively, PTQ can be applied to a pre-trained model
where weights are quantized after training a full precision
model \cite{vidya2021q}. The choice between PTQ and QAT depends on the accuracy and power requirements of the use case in hand \cite{liu2023quantization}.

\subsection{Neuromorphic Intermidiate Representation}
NIR provides a standardized set of model primitives designed as hybrid systems that combine continuous dynamics with discrete events \cite{pedersen2024neuromorphic}. By ignoring specific details of discretization and hardware, NIR accurately captures the computational model and bridges any gaps between the theoretical model and its implementation. It's goal is establishing a unified pipeline for deploying SNNs trained on various frameworks onto various platforms.

In this work, NIR was used to link software-based SNN models trained using SnnTorch with SpiNNaker2 deployment specifications. Model layers were represented as connected graph nodes, producing a NIR graph for each imported SNN model:
\textbf{This implementation introduces support for representing quantized parameters in NIR graphs},
highlighting NIR’s adaptability for different hardware needs.

\subsection{The SpiNNaker2 System}
SpiNNaker 2 is a multi-core digital neurmorphic chip \cite{gonzalez2024spinnaker2}. Each chip contains 152 processing elements (PEs) connected through a network-on-chip (NoC). Each PE includes an Arm M4F core, 128 KB of SRAM, and specialized accelerators for exponential functions, random number generation, and multiply-accumulate (MAC) operations. The chip includes a total of 19 MB on-chip SRAM, supplemented by 2 GB of external LPDDR4 memory. For our SNN models deployment on SpiNNaker 2, we utilize only the on-chip SRAM of a single chip, maximizing its capability to handle demanding tasks like gesture recognition by distributing computations across 147 of the 152 available PEs.

Each PE’s SRAM is divided into four 32 KB banks, with one bank generally reserved for program storage and the other three allocated for SNN weights and neuron state variables.
Each Arm core is assigned a population of neurons of the same type together with their incoming synapses.
All PEs are woken up synchronously in a regular interval (3.5 ms for our use case) to start the neuron and synapse processing for one time step.
If a neuron spikes, SpiNNaker multicast packets are sent via the NoC to other PEs. There, the spikes are buffered in a FIFO buffer in the SRAM and processed in the subsequent time step as described in \cite{hoppner2017dynamic}.

\subsection{DVS Gesture Recognition}
The DVS Gesture dataset collected event streams of multiple subjects performing a set of gestures from 10 predefined classes, 
such as clapping and air guitar, capturing dynamic temporal information \cite{amir2017low}. 
An 11th class is composed of 'other' gestures invented by the subjects. 
Each event represents a relative change in illumination, encoded with spatial (x, y) coordinates on a 128×128-pixel sensor and a timestamp at microsecond resolution. 
An example of such an event stream is visualized in Figure\ref{fig:full-pipeline}-A.
To fit within SpiNNaker 2’s memory limits, spatial downsampling was applied to reduce resolution to 32×32 pixels, and raw events were aggregated into frames by binning them in 1 ms intervals for both training and testing. 

To further enhance model accuracy, we applied data augmentation using Tonic’s transformation tools \cite{lenz_gregor_2021_5079802}. Specifically, we "denoised" the input data by filtering out events that occur outside a specified temporal window. Events were removed if no other events occurred within a 1-pixel spatial and 1 second time unit neighborhood.
\section{Methods}
\subsection{Training pipeline and architectures}\label{AA}
\paragraph{\textbf{Spiking Neural Networks}}  Both networks share an identical sequential topology, as outlined in Table~\ref{tab:model_architecture}, which presents the architecture of the P-SNN model. In the Q-SNN model, the convolutional and linear layers were replaced with quantized layers with a bitwidth of 8, respectively, while maintaining the same overall structure.

Our dataset includes both spatial and temporal features, prompting the use of convolutional layers to effectively extract spatial characteristics from frames constructed from the raw stream of events. Additionally, we employ LIF neurons in the hidden layers, which are suited for capturing the temporal aspects of gestures by encoding information through spikes.

\begin{table}[H]
    \centering
    \caption{NETWORK ARCHITECTURE SUMMARY FOR P-SNN model}
    \label{tab:model_architecture}
    \renewcommand{\arraystretch}{1.2} 
    \begin{tabular}{l c c c c c}
        & \textbf{Layer} & \textbf{Index} & \textbf{Kernel} & \textbf{Stride} & \textbf{Output Shape} \\[-0.5em]
        &                &                &                 &                 & \textbf{(C, H, W)} \\
        \cmidrule[0.5pt](lr){1-6} 
        \addlinespace[0.5pt] 
        \cmidrule[0.5pt](lr){1-6} 
            & Input &  &  & & $2 \times 32 \times 32$ \\
            & Conv2D & 0 & $5 \times 5$ & 2 & $16 \times 15 \times 15$ \\
            & LIF & 1 & &  & $16 \times 15 \times 15$ \\
            & Conv2D & 2 & $3 \times 3$ & 1 & $16 \times 15 \times 15$  \\
            & LIF & 3 & &  & $16 \times 15 \times 15$ \\
            & SumPool & 4 & $2 \times 2$ & $2 \times 2$ & $16 \times 7 \times 7$ \\
            & Conv2D & 5 & $3 \times 3$ & 1 & $8 \times 7 \times 7$ \\
            & LIF & 6 &  &  & $8 \times 7 \times 7$ \\
            & SumPool & 7 & $2 \times 2$ & $2 \times 2$ & $8 \times 3 \times 3$ \\
            & Flatten & 8 &  &  & 72 \\
            & Linear & 9 &  &  & 256 \\
            & LIF & 10 &  &  & 256 \\
            & Linear & 11 &  &  & 11 \\
            & LIF & 12 &  &  & 11 \\
            & Output &  &  &  & 11 \\
        \cmidrule[0.5pt](lr){1-6}
    \end{tabular}
\end{table}

\paragraph{\textbf{Slicing Method}} In our study, the accumulation method chosen for constructing frames from raw events is crucial, particularly for achieving high accuracy. We utilize a slicing technique that divides events into frames based on a fixed time window. All events within each time slice are summed up into a frame for each polarity.

Our findings indicate that a 1 ms time window provides optimal performance, as shown in Table~\ref{tab:SW_acc}. Further analysis revealed that while the number of events per recording varies significantly, the relatively consistent recording length in DVS gesture data makes time-window slicing particularly effective. This observation underscores the importance of selecting an appropriate time window for high-accuracy results. In terms of model size, both the P-SNN and Q-SNN models have 25,504 parameters, confirming that quantization only affects how the weights are stored. The P-SNN model uses full-precision 32-bit floating-point (FP32) representation, where each parameter consumes 4 bytes, while the Q-SNN model utilizes 8-bit integers (INT8), reducing the storage requirement to 1 byte per parameter \cite{jin2022f8net}. This reduction in bit width leads to significant memory savings, as the quantized model requires only 25\% of the memory used by the FP32 model, achieving efficient storage without altering the total parameter count \cite{jacob2018quantization}.
%
%
\begin{table}[H]
    \centering
    \caption{Comparison of P-SNN and Q-SNN On and Off chip performance for the DVS gesture dataset}
    \label{tab:SW_acc}
    \renewcommand{\arraystretch}{1.3}
    \begin{tabular}{l S[table-format=2.1] S[table-format=2.1]}
        \toprule
        \textbf{Metric} & {\textbf{P-SNN}} & {\textbf{Q-SNN}} \\
        \midrule
        Off-Chip Test Accuracy (\%) & 95.07 & 94.69 \\
        On-Chip Test Accuracy (\%) & 94.0 & 94.13 \\
        Parameters ($10^3$) & 25.5 & 25.5 \\
        Model Size (MB) & 0.17 & 0.04 \\
        \bottomrule
    \end{tabular}
\end{table}
\paragraph{\textbf{Setup}} To train our SNNs, we use the surrogate gradient method to approximate the derivatives of LIF neurons, addressing the non-differentiability of spikes with a fast sigmoid surrogate gradient \cite{neftci2019surrogate}-\cite{bellec2018long}.
We use a mean squared error (MSE) loss function for both the P-SNN and Q-SNN, with the Adam optimizer. We trained the models on a 16GB NVIDIA V100 GPU using Brevitas 0.10.2 for uniform quantization, snnTorch 0.9.1 for spiking neuron models \cite{eshraghian2023training}, Sinabs  2.0.0 \cite{Sheik_SINABS} for Sumpooling layers and PyTorch 2.2.0, all within Python 3.10.4, and 200 epochs for training.

All convolutional layers are set with padding and dilation of 1x1, and no biases are used in convolutional or fully connected layers. We save bias addition without observing significant impact on performance and it simplifies computation. Hyperparameters for each model are summarized in Table~\ref{tab:hyperparams}.

\begin{table*}[h!]
    \centering
    \caption{HYPERPARAMETER SUMMARY FOR THE TWO MODELS}
    \label{tab:hyperparams}
    \renewcommand{\arraystretch}{1.3}
    \begin{tabular}{ll S[table-format=2.0] S[table-format=0.2] S[table-format=0.4] S[table-format=1.2] ccc S[table-format=0.4]}
        \toprule
        \textbf{Model} & \textbf{Precision} & {\textbf{Batch size}} & {\textbf{Decay Rate $\beta$}} & {\textbf{Threshold $\theta$}} & {\textbf{Slope $k$}} & \textbf{Bias} & \textbf{Delay} & \textbf{Reset mechanism} & {\textbf{Learning Rate}} \\
        \midrule
        P-SNN & FP32 & 32 & 0.93 & 1 & 9.70 & False & 1 & Subtract & 0.0024 \\
        Q-SNN & INT8 & 32 & 0.93 & 1 & 9.50 & False & 1 & Subtract & 0.0030 \\
        \bottomrule
    \end{tabular}
\end{table*}

\subsection{Quantization Pipelines}

\paragraph{\textbf{PTQ}} We converted the P-SNN model to a NIR graph. Each model layer is represented as a graph node, with convolution layers storing parameters like the full precision weights, stride and padding while linear layers store only full precision weights. LIF neuron nodes include time constants \(\tau\), membrane resistance  \(r\), voltage leak, and thresholds, while pooling layers store kernel size, stride, and padding.

For deployment on the SpiNNaker2 chip, the floating point weights from the NIR graph (typical range: [-1.0, 1.0]) need to be converted to integer values between [-128, 127] while LIF neuron parameters and states can be implemented as 32-bit floating point in SpiNNaker2.
The neuron parameters from the NIR LIF model are translated to the SpiNNaker2 LIF implementation following the description in \cite{pedersen2024neuromorphic}.

Weights are scaled by a factor $\lambda_s$
\begin{align}
    w_\text{S2} = \lambda_s w_\text{NIR} \quad \label{eq:w_scale}
\end{align}

where for PTQ this factor is determined by analyzing the distribution of absolute weights for the NIR layer:
\begin{align}
{\lambda_s} &= \frac{127}{\left| W \right|_{\text{max}}} \\[0.5em]
\left| W \right|_{\text{max}} &= P_w(p) \quad 
\end{align}
Here, we use the percentile function $P_w(p)$, which calculates the \(p-th\) percentile of the incoming absolute weights for each neuron layer. A percentile value of $p=100$ means that the maximum absolute NIR weight will be scaled to an absolute weight of 127 on SpiNNaker2. Yet, in case of outliers in the weight distribution this \emph{max scaling} may lead to a severe drop in weight precision due to the quantization.
Hence, we experimented with various percentiles, from the 100th and 99th to lower values, as detailed in Section~\ref{sec:results}.

In order to retain the same spiking behaviour, the LIF firing thresholds ${\Gamma}$ were scaled accordingly: 
\begin{align}
    \Gamma_\text{S} = \lambda_s \Gamma \quad \label{eq:w_scale}
\end{align}

This process results in 8-bit weights and scaled firing thresholds compatible with the chip.

\paragraph{\textbf{QAT}} The generated NIR graph for the Q-SNN model retains the same overall structure as its predecessor, the P-SNN model. This includes maintaining consistency in node types, quantities, and layer indices. However, a key distinction lies in the weight nodes. In the Q-SNN model, quantized convolution and linear layers are employed, which store both full-precision and 8-bit weights, along with their corresponding scaling factors. During training, Brevitas utilizes these scaling factors ${S}$ to derive the 8-bit weights ${w_q}$ from the full-precision weights ${w_r}$ \cite{nagel2021white}-\cite{kim2022integer}, ensuring precise quantization and compatibility.

\begin{align}
    w_\text{q} = \lambda_q w_\text{r} \quad 
\end{align}

\begin{align}
    \lambda_q = \frac{1}{S} \quad \label{eq:lambda_q}
\end{align}

In our work, we also store these scaling factors within the NIR graph as node metadata to enable adaptive threshold scaling in subsequent stages. The QAT pipeline with Adaptive LIF Threshold Scaling is outlined in Algorithm~\ref{alg:lif_scaling}.

\begin{algorithm}[H]
\caption{Adaptive LIF Threshold Scaling in QAT Pipeline}
\label{alg:lif_scaling}
\begin{algorithmic}[1]
    \STATE \textbf{Parameters:} Scaling factors $S$, $w_q$, and $w_r$ weights for each quantized layer.
    \STATE Initialize empty list $\Gamma_s$.
    
    \FOR{$i = 1$ to $N$ (number of layers)}
        \STATE Extract scaling factor $S_i$.
        \STATE Apply quantized layer with $w_q$ weights:
        \STATE \quad $L_{i-1} \gets \text{QuantizedLayer}(x, w_q, \text{type})$
        \STATE Compute scaled LIF threshold for layer $i$:
        \STATE \quad $\Gamma_i \gets \frac{\Gamma_i}{S_i}$
        \STATE Append $\Gamma_i$ to $\Gamma_s$.
        \STATE Apply LIF layer with the scaled threshold:
        \STATE \quad $x \gets \text{LIF}(L_i, \Gamma_i)$
    \ENDFOR
    
    \STATE \textbf{Output:} $\Gamma_s$ (Scaled thresholds for all LIF layers).
\end{algorithmic}
\end{algorithm}

It dynamically adjusts the LIF thresholds for each quantized layer. The process begins by extracting predefined scaling factors and scaling the LIF threshold of each layer based on the scaling factor of the preceding weight layer, effectively accounting for the impact of quantization on threshold values. These scaled thresholds are then applied to the respective LIF layers. The final result is 8-bit weights and adaptively scaled firing thresholds.
Finally, when converting the quantized NIR model to py-spinnaker2, the weight scaling as applied in PTQ is switched off ($\lambda_s=1$), and the quantized weights are used directly.

\subsection{SpiNNaker2 Implementation} \label{sec:s2}
For the implementation on the chip we use the software py-spinnaker2 \cite{vogginger2023pyspinnaker2} that provides a light-weight Python interface for running experiments on a single-chip SpiNNaker2 test board. It uses 8-bit signed synapse weights and 32-bit floating-point
numbers for neuron parameters and state variables respectively. The API for defining SNN models is inspired by pyNN \cite{davison2009pynn}.

To integrate our models with SpiNNaker2 hardware, the NIR graphs were converted into a format compatible with the SpiNNaker2 network. After completing the quantization step either through the PTQ for the P-SNN or QAT for the Q-SNN, we ensured that both the weights and firing thresholds fit within the chip’s dynamic range. Next, we translated each LIF layer from our model into a "population" within the SpiNNaker2 network. A population represents a group of neurons following the same neuron model, which, in our case, is LIF. This model records input and output spikes at the specific time steps they occur, starting from a time step of 0.

 Subsequently, each layer in the model positioned between two consecutive LIF layers (such as convolution-only, sum-pooling followed by convolution, sum-pooling followed by flatten and linear layers, or linear-only layers) was converted into a "projection" linking these consecutive LIF populations. A projection consists of a list of synapses between two populations, with parameters defining the pre-synaptic and post-synaptic populations, as well as a list of synaptic connections. A summary of the conversion parameters for the two NIR graphs is shown in Table~\ref{tab:conversion}.
\begin{table*}[h!]
    \centering
    \caption{NIR to SpiNNaker2 conversion hyperparameter SUMMARY}
    \label{tab:conversion}
    \renewcommand{\arraystretch}{1.3}
    \begin{tabular}{llccccc}
        \toprule
        \textbf{Model} & \textbf{Recordable} & \textbf{Delay} & \textbf{Scale Thresholds} & \textbf{Weight Percentile} & \textbf{Reset Mechanism} & \textbf{Integrator Mechanism} \\
        \midrule
        P-SNN & Spikes & 1 & True & [90, 100] & Subtract & Euler-Forward \\
        Q-SNN & Spikes & 1 & True & $\boldsymbol{\times}$ & Subtract & Euler-Forward\\
        \bottomrule
    \end{tabular}
\end{table*}

\begin{figure}[h]
    \centering
    \includegraphics[width=0.5\textwidth]{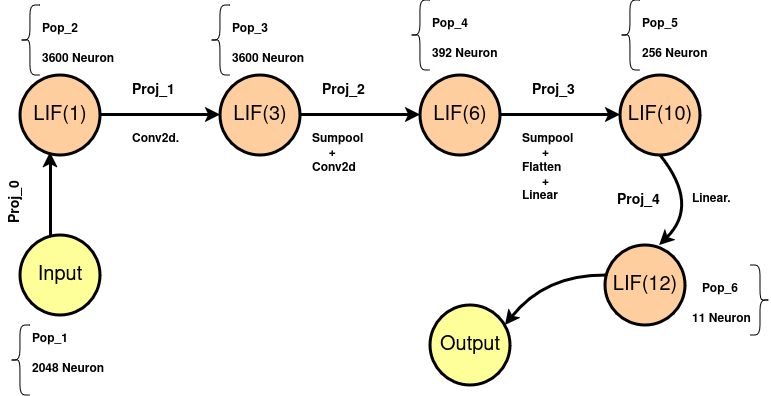} 
    \caption{SpiNNaker2 Network Architecture: Visualization of LIF neuron populations, indicating the number of neurons in each population and the synaptic projection connecting them. 
}
    \label{fig:sp2_network}
\end{figure}

In total, our SpiNNaker2 network consists of six populations: five LIF populations and one input population, connected by five projections and an output that enables recording on pre-node as depicted in Figure~\ref{fig:sp2_network}. Eventually, from each NIR graph we generate a separate network.

A manual partitioning was applied in order to map the network on the SpiNNaker2 chip.
This step is needed as the maximum number of neurons that can be implemented on single PE is limited by the SRAM, as reported in Table~\ref{tab:max_atome}. This maximum mainly depends on the complexity of the neuron models and how incoming synapses are represented.
If a population exceeds this number, it is automatically mapped onto multiple PEs with each PE holding only a slice of the population.
\begin{table}[H]
    \centering
    \caption{Constraints of the SpiNNaker2 PE}
    \label{tab:max_atome}
    \renewcommand{\arraystretch}{1.3}
    \begin{tabular}{lc}
        \toprule
        \textbf{Neuron Model} & \textbf{Max Neurons} \\
        \midrule
        LIF\_Conv2d & 1024 \\
        LIF\_Neuron & 250 \\
        Spike\_List & 500 \\
        \bottomrule
    \end{tabular}
\end{table}
Yet, even if a population with less than the maximum neurons is mapped to one PE, it can happen that the SRAM is not large enough for storing all synapses and for recording all spikes. This may happen especially for fully-connected layers and when spikes are recorded for long simulation times.
To avoid these memory limitations of the current software which does not use the DRAM, we manually reduce the maximum number of neurons per PE such that a population is distributed across more PEs as shown in Table~\ref{tab:mapping}.
\begin{table}[H]
    \centering
    \caption{Partitioning per PE of the P-SNN and Q-SNN networks on SpiNNaker2 
    }
    \label{tab:mapping}
    \renewcommand{\arraystretch}{1.3}
    \begin{tabular}{lc}
        \toprule
        \textbf{Population} & \textbf{Max Neurons} \\
        \midrule
        LIF(1) & 900 \\
        LIF(3) & 900 \\
        LIF(6) & 980 \\
        LIF(10) & 16 \\
        input & 17 \\
        \bottomrule
    \end{tabular}
\end{table}

Additionally, to further address these memory constraints, we limited the simulation to approximately 600 timesteps (roughly 600 milliseconds) of each gesture, compared to the full gesture duration of around 6 seconds. This adjustment ensured sufficient memory for storing synapses and spikes while staying within the chip’s SRAM limitations.

\section{Experiment and evaluation} \label{sec:results}
\begin{table*}[!htb]
    \centering
    \caption{Comparison of the results with the state of the art embedded and spiking neural network for gesture recognition implementation}
    \label{tab:SOTA}
    \renewcommand{\arraystretch}{1.2} 
    \begin{tabular}{l c c c c c c c c c}
        & \textbf{} & \textbf{\cite{amir2017low}} & \textbf{\cite{eshraghian2022navigating}} & \textbf{\cite{venkatesh2024squat}} & \textbf{\cite{innocenti2021temporal}} & \textbf{\cite{yao2024spike}} & \textbf{\cite{massa2020efficient}} & \multicolumn{2}{c}{\textbf{This work}} \\
        \cmidrule[0.5pt](lr){1-10}
        & & & & & & & & \textbf{P-SNN} & \textbf{Q-SNN} \\
        \cmidrule[0.5pt](lr){1-10}
        & Input format & Events  & Events & Events & Events & Events & Frames & Events & Events \\
        \cmidrule[0.5pt](lr){1-10}
        & Neural network architecture & SNN & SNN & SNN & 3D CNN & SNN & SNN & PTQ-SNN & QAT-SNN \\
        \cmidrule[0.5pt](lr){1-10}
        & Inference hardware & TrueNorth & GPU & GPU & GPU & Speck & Loihi & SpiNNaker2 & SpiNNaker2 \\
        \cmidrule[0.5pt](lr){1-10}
        & Training Method & CNN-to-SNN & Surrogate & Surrogate & - & Surrogate  &  & Surrogate  & Surrogate  \\
        \cmidrule[0.5pt](lr){1-10}
        & Quantization Method & Deterministic rounding & QAT  & QAT  & - &  & - &  PTQ  & QAT \\
        \cmidrule[0.5pt](lr){1-10}
        & Weight bitwidth & Ternary & 32  & 8  & 8  & -  & 9  & 8  & 8  \\
        \cmidrule[0.5pt](lr){1-10}
        & Energy per Inference (mJ) & 18.8  & - & - & & & - & 459 & 459 \\
        \cmidrule[0.5pt](lr){1-10}
        & Power (mW) & 44.5  & - & - & - & 3.8  & 137 & - & - \\
        \cmidrule[0.5pt](lr){1-10}
        & Model Size (MB) & 38 & - & - & - & - & - & 0.17 & 0.04 \\
        \cmidrule[0.5pt](lr){1-10}
        & Statistical Accuracy (\%) & 94.6  & 93.05  & 83.97  & 99.6  & 90.0 & 89.64 & 94.0 & 94.13 \\
        \cmidrule[0.5pt](lr){1-10}
    \end{tabular}
\end{table*}

\subsection{Comparison with Prior Work}
The comparison results in Table~\ref{tab:SOTA} highlight how our models perform relative to other state-of-the-art full-precision and quantized SNN implementations on the DVS Gesture dataset, including neuromorphic hardware deployment. 

To ensure a fair comparison, models that utilize GPUs for inference are benchmarked against the results of our models on GPU, as shown in Table~\ref{tab:SW_acc}.

Our full-precision SNN model achieves a high performance improvement over the model presented in \cite{eshraghian2022navigating}. Additionally, our 8-bit quantized model surpasses the results of the 8-bit model in \cite{venkatesh2024squat}, which employed QAT for weights only and used a surrogate gradient method for training.

In terms of neuromorphic hardware deployment, our P-SNN and Q-SNN models running on SpiNNaker2 demonstrate superior inference performance compared to the results achieved on the Loihi chip, as reported in \cite{massa2020efficient} and also the Speck chip in \cite{yao2024spike}.

\subsection{Quantization-Accuracy Drop Trade-off}

\begin{figure}[H]
    \centering
    \includegraphics[width=0.5\textwidth]{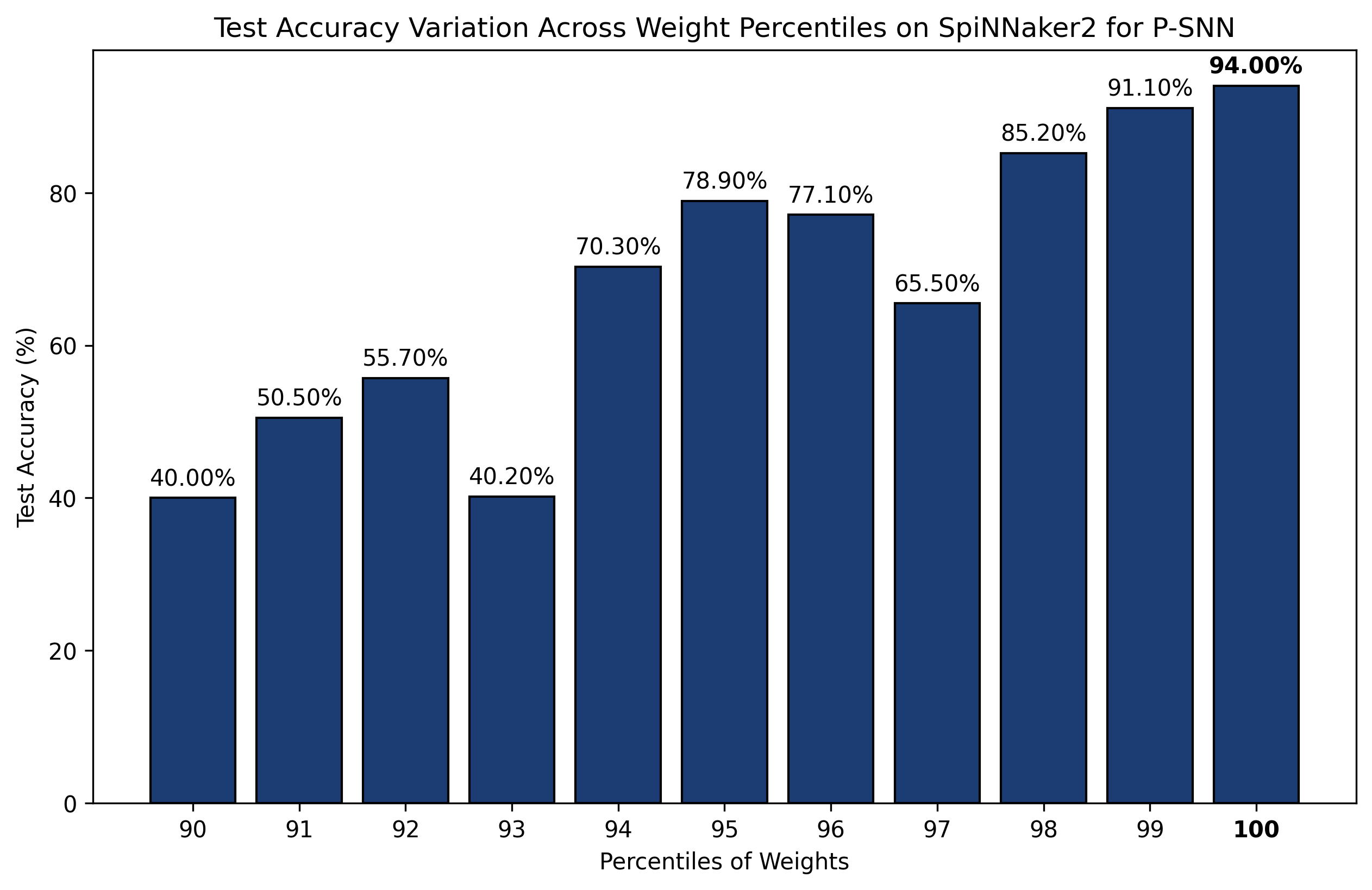} 
    \caption{Profiling the on-chip classification accuracy for different percentile values. The highest accuracy is achieved at 100th percentile of weights.}
    \label{fig:ptq_p}
\end{figure}

For the P-SNN model, the baseline accuracy is \SI{95.07}{\percent}. After quantization with PTQ at different percentiles, the on-chip accuracy drops to \SI{94.0}{\percent}, which was achieved at the 100th percentile, as shown in Figure  \ref{fig:ptq_p}. 

This means that the PTQ pipeline kept the accuracy degradation within \textbf{1.07}\,\textbf{\%} of the baseline.

On the other hand, the Q-SNN model, that was trained using QAT with careful fine-tuning step during training, achieves a baseline accuracy of \SI{94.69}{\percent}
For the inference on SpiNNaker2 the accuracy is \SI{94.13}{\percent}, resulting in a degradation of \textbf{0.56}\,\textbf{\%} from the baseline.

To address the baseline differences between the two pipelines, Quantization Aware Fine-Tuning (QAF) was employed as a software optimization strategy for the Q-SNN model. This approach ensures the model adapts effectively to quantization-induced changes, preserving performance. 
Additionally, the performance degradation in the QAT pipeline is significantly lower than that of PTQ. This suggests that the QAT pipeline is better suited for tasks on SpiNNaker2 than PTQ.

The slight accuracy drops observed for both SNN models on-chip are likely due to noise introduced during the quantization process, which is an inherent trade-off for achieving efficient hardware deployment. Additionally, this minor decrease in performance can be attributed to the fact that not all timesteps are simulated on the SpiNNaker2, as discussed in subsection~\ref{sec:s2}. This limitation stems from the current software stack but will be resolved once DRAM integration is completed in py-spinnaker2. Another possible contributing factor could be differences between the software implementation in snntorch and the hardware behavior of SpiNNaker2, 
particularly when exactly the membrane voltage reset is  applied in each environment.

\subsection{Energy Consumption}


The SpiNNaker2 chip supports dynamic voltage and frequency scaling (DVFS) per PE\cite{hoppner2017dynamic} allowing to switch between a high-performance and a low-power mode. Here, the high performance level with 300 MHz clock frequency and 0.8 V supply voltage is used.
The energy consumption of both the P-SNN and Q-SNN on SpiNNaker2 are given in Table~\ref{tab:SOTA}, representing the average energy per gesture. Per 1 ms frame the inference energy is 0.765 \, \text{mJ}.

\section{Conclusion}
In this paper, we propose an efficient method for deploying Deep Spiking Neural Networks (DSNNs) on the SpiNNaker2 neuromorphic chip for the DVS gesture recognition task using the neuromorphic intermediate representation (NIR). After conducting a comparative study of two quantization pipelines, post training quantization (PTQ) and quantization aware training (QAT), our results demonstrate that QAT is better suited for accurate inference on neuromorphic processors with minimal performance degradation.

To address the quantization effects and accuracy drops encountered during SNN inference on hardware with stringent memory constraints, we promoted adding support for QAT in NIR. This enhancement will pave the way for future work to optimize accuracy and performance further.

Additionally, we plan to utilize the LPDDR4 memory on the SpiNNaker2 chip to store input spike streams of gestures, which will accelerate spike processing in the input layer. This approach aims to reduce the system ticks per second, making the application more suitable for real-time use. Finally, our two SNN models can leverage SpiNNaker2’s scalable design to distribute workloads efficiently across multiple chips \cite{nazeer2024language}, enabling enhanced performance in distributed computing scenarios.

\section*{Acknowledgment} 
The authors thank Matthias Jobst for the insightful discussions and feedback regarding NIR to SpiNNaker2 implementation and fine-tuning in QAT.
This work is partly funded by the European Union within the programme Horizon Europe under grant agreement no.~101120727 (PRIMI).
Mark Sch\"{o}ne is fully funded by the Bosch Research Foundation. Christian Mayr is funded by the German Research Foundation (DFG, Deutsche Forschungsgemeinschaft) as
part of Germany’s Excellence Strategy – EXC 2050/1 – Project ID 390696704 – Cluster of Excellence “Centre for Tactile
Internet with Human-in-the-Loop” (CeTI) of Technische Universit\"{a}t Dresden.

The
authors gratefully acknowledge the computing time made
available to them on the high-performance computer at
the NHR Center of TU Dresden. This center is jointly
supported by the Federal Ministry of Education and
Research and the state governments participating in the
NHR (www.nhr-verein.de/unsere-partner)


\bibliographystyle{IEEEtran}
\bibliography{references}

\begin{thebibliography}{10}
\providecommand{\url}[1]{#1}
\csname url@samestyle\endcsname
\providecommand{\newblock}{\relax}
\providecommand{\bibinfo}[2]{#2}
\providecommand{\BIBentrySTDinterwordspacing}{\spaceskip=0pt\relax}
\providecommand{\BIBentryALTinterwordstretchfactor}{4}
\providecommand{\BIBentryALTinterwordspacing}{\spaceskip=\fontdimen2\font plus
\BIBentryALTinterwordstretchfactor\fontdimen3\font minus \fontdimen4\font\relax}
\providecommand{\BIBforeignlanguage}[2]{{%
\expandafter\ifx\csname l@#1\endcsname\relax
\typeout{** WARNING: IEEEtran.bst: No hyphenation pattern has been}%
\typeout{** loaded for the language `#1'. Using the pattern for}%
\typeout{** the default language instead.}%
\else
\language=\csname l@#1\endcsname
\fi
#2}}
\providecommand{\BIBdecl}{\relax}
\BIBdecl

\bibitem{pfeiffer2018deep}
M.~Pfeiffer and T.~Pfeil, ``Deep learning with spiking neurons: Opportunities and challenges,'' \emph{Frontiers in neuroscience}, vol.~12, p. 409662, 2018.

\bibitem{weng2021neural}
O.~Weng, ``Neural network quantization for efficient inference: A survey,'' \emph{arXiv preprint arXiv:2112.06126}, 2021.

\bibitem{vidya2020fspinn}
R.~Vidya Wicaksana~Putra and M.~Shafique, ``Fspinn: An optimization framework for memory-and energy-efficient spiking neural networks,'' \emph{arXiv e-prints}, pp. arXiv--2007, 2020.

\bibitem{rathi2018stdp}
N.~Rathi, P.~Panda, and K.~Roy, ``Stdp-based pruning of connections and weight quantization in spiking neural networks for energy-efficient recognition,'' \emph{IEEE Transactions on Computer-Aided Design of Integrated Circuits and Systems}, vol.~38, no.~4, pp. 668--677, 2018.

\bibitem{sen2017approximate}
S.~Sen, S.~Venkataramani, and A.~Raghunathan, ``Approximate computing for spiking neural networks,'' in \emph{Design, Automation \& Test in Europe Conference \& Exhibition (DATE), 2017}.\hskip 1em plus 0.5em minus 0.4em\relax IEEE, 2017, pp. 193--198.

\bibitem{amir2017low}
A.~Amir, B.~Taba, D.~Berg, T.~Melano, J.~McKinstry, C.~Di~Nolfo, T.~Nayak, A.~Andreopoulos, G.~Garreau, M.~Mendoza \emph{et~al.}, ``A low power, fully event-based gesture recognition system,'' in \emph{Proceedings of the IEEE conference on computer vision and pattern recognition}, 2017, pp. 7243--7252.

\bibitem{eshraghian2023training}
J.~K. Eshraghian, M.~Ward, E.~O. Neftci, X.~Wang, G.~Lenz, G.~Dwivedi, M.~Bennamoun, D.~S. Jeong, and W.~D. Lu, ``Training spiking neural networks using lessons from deep learning,'' \emph{Proceedings of the IEEE}, vol. 111, no.~9, pp. 1016--1054, 2023.

\bibitem{pedersen2024neuromorphic}
J.~E. Pedersen, S.~Abreu, M.~Jobst, G.~Lenz, V.~Fra, F.~C. Bauer, D.~R. Muir, P.~Zhou, B.~Vogginger, K.~Heckel \emph{et~al.}, ``Neuromorphic intermediate representation: A unified instruction set for interoperable brain-inspired computing,'' \emph{Nature Communications}, vol.~15, no.~1, p. 8122, 2024.

\bibitem{vogginger2023pyspinnaker2}
\BIBentryALTinterwordspacing
B.~Vogginger, F.~Kelber, M.~Jobst, Y.~Yan, P.~Gerhards, M.~Weih, and M.~Akl, ``py-spinnaker2,'' Nov. 2023. [Online]. Available: \url{https://doi.org/10.5281/zenodo.10202110}
\BIBentrySTDinterwordspacing

\bibitem{vidya2021q}
R.~Vidya Wicaksana~Putra and M.~Shafique, ``Q-spinn: A framework for quantizing spiking neural networks,'' \emph{arXiv e-prints}, pp. arXiv--2107, 2021.

\bibitem{eshraghian2022fine}
J.~K. Eshraghian and W.~D. Lu, ``The fine line between dead neurons and sparsity in binarized spiking neural networks,'' \emph{arXiv preprint arXiv:2201.11915}, 2022.

\bibitem{lava-dl}
``lava-dl,'' \url{ https://github.com/lava-nc/lava-dl }.

\bibitem{massa2020efficient}
R.~Massa, A.~Marchisio, M.~Martina, and M.~Shafique, ``An efficient spiking neural network for recognizing gestures with a dvs camera on the loihi neuromorphic processor,'' in \emph{2020 International Joint Conference on Neural Networks (IJCNN)}.\hskip 1em plus 0.5em minus 0.4em\relax IEEE, 2020, pp. 1--9.

\bibitem{eshraghian2022navigating}
J.~K. Eshraghian, C.~Lammie, M.~R. Azghadi, and W.~D. Lu, ``Navigating local minima in quantized spiking neural networks,'' in \emph{2022 IEEE 4th International Conference on Artificial Intelligence Circuits and Systems (AICAS)}.\hskip 1em plus 0.5em minus 0.4em\relax IEEE, 2022, pp. 352--355.

\bibitem{severa2018whetstone}
W.~M. Severa, C.~M. Vineyard, R.~Dellana, and J.~B. Aimone, ``Whetstone: An accessible platform-independent method for training spiking deep neural networks for neuromorphic processors.'' Sandia National Lab.(SNL-NM), Albuquerque, NM (United States), Tech. Rep., 2018.

\bibitem{venkatesh2024squat}
S.~Venkatesh, R.~Marinescu, and J.~K. Eshraghian, ``Squat: stateful quantization-aware training in recurrent spiking neural networks,'' in \emph{2024 Neuro Inspired Computational Elements Conference (NICE)}.\hskip 1em plus 0.5em minus 0.4em\relax IEEE, 2024, pp. 1--10.

\bibitem{bengio2013estimating}
Y.~Bengio, ``Estimating or propagating gradients through stochastic neurons,'' \emph{arXiv preprint arXiv:1305.2982}, 2013.

\bibitem{fan2022training}
T.-H. Fan, T.-C. Chi, A.~I. Rudnicky, and P.~J. Ramadge, ``Training discrete deep generative models via gapped straight-through estimator,'' in \emph{International Conference on Machine Learning}.\hskip 1em plus 0.5em minus 0.4em\relax PMLR, 2022, pp. 6059--6073.

\bibitem{liu2023quantization}
S.~Liu, N.~Mohammadi, and Y.~Yi, ``Quantization-aware training of spiking neural networks for energy-efficient spectrum sensing on loihi chip,'' \emph{IEEE Transactions on Green Communications and Networking}, vol.~8, no.~2, pp. 827--838, 2023.

\bibitem{gonzalez2024spinnaker2}
H.~A. Gonzalez, J.~Huang, F.~Kelber, K.~K. Nazeer, T.~Langer, C.~Liu, M.~Lohrmann, A.~Rostami, M.~Sch{\"o}ne, B.~Vogginger \emph{et~al.}, ``Spinnaker2: A large-scale neuromorphic system for event-based and asynchronous machine learning,'' \emph{arXiv preprint arXiv:2401.04491}, 2024.

\bibitem{hoppner2017dynamic}
S.~H{\"o}ppner, Y.~Yan, B.~Vogginger, A.~Dixius, J.~Partzsch, F.~Neum{\"a}rker, S.~Hartmann, S.~Schiefer, S.~Scholze, G.~Ellguth \emph{et~al.}, ``Dynamic voltage and frequency scaling for neuromorphic many-core systems,'' in \emph{2017 IEEE International Symposium on Circuits and Systems (ISCAS)}.\hskip 1em plus 0.5em minus 0.4em\relax IEEE, 2017, pp. 1--4.

\bibitem{lenz_gregor_2021_5079802}
\BIBentryALTinterwordspacing
G.~Lenz, K.~Chaney, S.~B. Shrestha, O.~Oubari, S.~Picaud, and G.~Zarrella, ``Tonic: event-based datasets and transformations.'' Jul. 2021, {Documentation available under https://tonic.readthedocs.io}. [Online]. Available: \url{https://doi.org/10.5281/zenodo.5079802}
\BIBentrySTDinterwordspacing

\bibitem{jin2022f8net}
Q.~Jin, J.~Ren, R.~Zhuang, S.~Hanumante, Z.~Li, Z.~Chen, Y.~Wang, K.~Yang, and S.~Tulyakov, ``F8net: Fixed-point 8-bit only multiplication for network quantization,'' \emph{arXiv preprint arXiv:2202.05239}, 2022.

\bibitem{jacob2018quantization}
B.~Jacob, S.~Kligys, B.~Chen, M.~Zhu, M.~Tang, A.~Howard, H.~Adam, and D.~Kalenichenko, ``Quantization and training of neural networks for efficient integer-arithmetic-only inference,'' in \emph{Proceedings of the IEEE conference on computer vision and pattern recognition}, 2018, pp. 2704--2713.

\bibitem{neftci2019surrogate}
E.~O. Neftci, H.~Mostafa, and F.~Zenke, ``Surrogate gradient learning in spiking neural networks: Bringing the power of gradient-based optimization to spiking neural networks,'' \emph{IEEE Signal Processing Magazine}, vol.~36, no.~6, pp. 51--63, 2019.

\bibitem{bellec2018long}
G.~Bellec, D.~Salaj, A.~Subramoney, R.~Legenstein, and W.~Maass, ``Long short-term memory and learning-to-learn in networks of spiking neurons,'' \emph{Advances in neural information processing systems}, vol.~31, 2018.

\bibitem{Sheik_SINABS}
S.~Sheik, G.~Lenz, F.~Bauer, and N.~Kuepelioglu, ``{SINABS: A simple Pytorch based SNN library specialised for Speck},'' 2023, https://github.com/synsense/sinabs.

\bibitem{nagel2021white}
M.~Nagel, M.~Fournarakis, R.~A. Amjad, Y.~Bondarenko, M.~Van~Baalen, and T.~Blankevoort, ``A white paper on neural network quantization,'' \emph{arXiv preprint arXiv:2106.08295}, 2021.

\bibitem{kim2022integer}
S.~Kim, A.~Gholami, Z.~Yao, N.~Lee, P.~Wang, A.~Nrusimha, B.~Zhai, T.~Gao, M.~W. Mahoney, and K.~Keutzer, ``Integer-only zero-shot quantization for efficient speech recognition,'' in \emph{ICASSP 2022-2022 IEEE International Conference on Acoustics, Speech and Signal Processing (ICASSP)}.\hskip 1em plus 0.5em minus 0.4em\relax IEEE, 2022, pp. 4288--4292.

\bibitem{davison2009pynn}
A.~P. Davison, D.~Br{\"u}derle, J.~M. Eppler, J.~Kremkow, E.~Muller, D.~Pecevski, L.~Perrinet, and P.~Yger, ``Pynn: a common interface for neuronal network simulators,'' \emph{Frontiers in neuroinformatics}, vol.~2, p. 388, 2009.

\bibitem{innocenti2021temporal}
S.~U. Innocenti, F.~Becattini, F.~Pernici, and A.~Del~Bimbo, ``Temporal binary representation for event-based action recognition,'' in \emph{2020 25th International Conference on Pattern Recognition (ICPR)}.\hskip 1em plus 0.5em minus 0.4em\relax IEEE, 2021, pp. 10\,426--10\,432.

\bibitem{yao2024spike}
M.~Yao, O.~Richter, G.~Zhao, N.~Qiao, Y.~Xing, D.~Wang, T.~Hu, W.~Fang, T.~Demirci, M.~De~Marchi \emph{et~al.}, ``Spike-based dynamic computing with asynchronous sensing-computing neuromorphic chip,'' \emph{Nature Communications}, vol.~15, no.~1, p. 4464, 2024.

\bibitem{nazeer2024language}
K.~K. Nazeer, M.~Sch{\"o}ne, R.~Mukherji, B.~Vogginger, C.~Mayr, D.~Kappel, and A.~Subramoney, ``Language modeling on a spinnaker2 neuromorphic chip,'' in \emph{2024 IEEE 6th International Conference on AI Circuits and Systems (AICAS)}.\hskip 1em plus 0.5em minus 0.4em\relax IEEE, 2024, pp. 492--496.

\end{thebibliography}

\end{document}